\documentclass[conference]{IEEEtran}
\IEEEoverridecommandlockouts

\usepackage{cite}
\usepackage{amsmath,amssymb,amsfonts}
\usepackage{pifont}
\usepackage{graphicx}
\usepackage{textcomp}
\usepackage{xcolor}
\def\BibTeX{{\rm B\kern-.05em{\sc i\kern-.025em b}\kern-.08em
    T\kern-.1667em\lower.7ex\hbox{E}\kern-.125emX}}

\usepackage{times} 
\usepackage{helvet}  
\usepackage{courier} 
\usepackage[hyphens]{url}  
\usepackage{caption}
\usepackage{algorithm}
\usepackage[noend]{algpseudocode}
\usepackage{newfloat}
\usepackage{listings}
\usepackage{comment}
\usepackage{booktabs}
\usepackage{subcaption}
\usepackage{multirow}
\usepackage{balance}

\usepackage{circledsteps}

\begin{document}

\title{ICE-Pruning: An \underline{I}terative \underline{C}ost-\underline{E}fficient Pruning Pipeline for Deep Neural Networks

}

\begin{comment}
%\author{\IEEEauthorblockN{Anonymous Authors}}
\author{\IEEEauthorblockN{Wenhao Hu}
\IEEEauthorblockA{\textit{School of Computing Science} \\
\textit{University of Glasgow}\\
Glasgow, United Kingdom \\
2692597H@student.gla.ac.uk}
\and
\IEEEauthorblockN{Paul Henderson}
\IEEEauthorblockA{\textit{School of Computing Science} \\
\textit{University of Glasgow}\\
Glasgow, United Kingdom \\
Paul.Henderson@glasgow.ac.uk}
\and
\IEEEauthorblockN{José Cano}
\IEEEauthorblockA{\textit{School of Computing Science} \\
\textit{University of Glasgow}\\
Glasgow, United Kingdom \\
Jose.CanoReyes@glasgow.ac.uk}
}
\end{comment}

\author{Wenhao Hu, Paul Henderson, Jos\'e Cano \\
\emph{School of Computing Science, University of Glasgow, Scotland, UK}
}

\maketitle

%***************************************************************************************************

\begin{abstract}

Pruning is a widely used method for compressing Deep Neural Networks (DNNs), where less relevant parameters are removed from a DNN model to reduce its size. However, removing parameters reduces model accuracy, so pruning is typically combined with fine-tuning, and sometimes other operations such as rewinding weights, to recover accuracy.
A common approach is to repeatedly prune and then fine-tune, with increasing amounts of model parameters being removed in each step.
While straightforward to implement, pruning pipelines that follow this approach are computationally expensive due to the need for repeated fine-tuning.

In this paper we propose ICE-Pruning, an iterative pruning pipeline for DNNs that significantly decreases the time required for pruning by reducing the overall cost of fine-tuning, while maintaining a similar accuracy to existing pruning pipelines. 
ICE-Pruning is based on three main components: i) an automatic mechanism to determine after which pruning steps fine-tuning should be performed; 
ii) a freezing strategy for faster fine-tuning in each pruning step; 
and iii) a custom pruning-aware learning rate scheduler to further improve the accuracy of each pruning step and reduce the overall time consumption. 
We also propose an efficient auto-tuning stage for the hyperparameters (e.g., freezing percentage) introduced by the three components.
We evaluate ICE-Pruning on several DNN models and datasets, showing that it can accelerate pruning by up to 9.61$\times$. 
Code is available at \texttt{https://github.com/gicLAB/ICE-Pruning}

\begin{comment}
Pruning is one of the main compression methods for Deep Neural Networks (DNNs), where less relevant parameters are removed from a DNN model to reduce its memory footprint; however, it can reduce the model accuracy. Normally, pruning is performed within a pipeline that combines pruning, fine-tuning, and maybe other operations such as rewinding weights.
To get better final accuracy, as a naive pipeline, pruning is often performed iteratively with increasing amounts of parameters being removed in each step, and fine-tuning is applied to the remaining parameters. 
Based on that, other more sophisticated DNN pruning pipelines are created. However, despite the final model accuracy, existing pipelines are usually computationally expensive, especially for large DNNs and datasets.

Motivated by this, in this paper we propose ICE-Pruning, an iterative pruning pipeline for DNNs that significantly reduces the time required for DNN pruning while maintaining a similar final accuracy to existing pipelines. 
ICE-Pruning is based on three main components: i) an automatic mechanism to determine after which pruning steps fine-tuning should be performed; ii) a freezing strategy for faster fine-tuning in each pruning step; and iii) a custom, pruning-aware learning rate scheduler whose hyperparameters can be automatically configured.
We evaluate ICE-Pruning with three different DNN models and two datasets. The results show that it can accelerate pruning by up to 9.61$\times$ while maintaining accuracy.
\end{comment}

\end{abstract}

%***************************************************************************************************


%***************************************************************************************************

\section{Introduction}
\label{sec:intro}

Pruning is a widely used method for compressing Deep Neural Network (DNN) models, reducing their size and making them more suitable for deployment in resource-constrained edge devices such as mobile phones, drones, etc~\cite{survey,gibson_dlas_2025}.
It typically involves removing less important parameters from a pre-trained DNN model, followed by fine-tuning the remaining parameters on the training dataset to compensate for any potential loss in accuracy~\cite{LRF,entropy,ThinNet}.
The pruning process of a DNN is often formulated as a \textit{pruning-fine-tuning pipeline}, where inputs are a pre-trained DNN model, a dataset, and a pruning ratio; and the output is a pruned model that ideally achieves similar accuracy to the original model.

One popular design for the pruning-fine-tuning pipeline is \textit{iterative pruning-fine-tuning}.
Here the DNN model is repeatedly pruned, with fine-tuning performed after each pruning step; this process continues until a target compression ratio is reached~\cite{l1, entropy,LRF}.
However, the repeated fine-tuning operations in such pipelines are prohibitively time-consuming when the dataset and/or DNN model are large.
More sophisticated methods adapt this basic pipeline in different ways. For example, AAP~\cite{AAP} sometimes rewinds the weights of the pruned model to an earlier state and resumes pruning from there.
Unfortunately, this further increases the time consumption of the pruning process.

To address the slowness of existing iterative pruning pipelines, we propose \textit{ICE-Pruning}, a novel and automatically configured Iterative Cost-Efficient pruning pipeline that significantly reduces the pruning time while providing similar accuracy to existing pipelines, especially for high-redundancy DNN models.  
ICE-Pruning is based on the idea that, due to the high parameter redundancy present in most DNN models, not all the pruning steps need to be followed by fine-tuning to recover accuracy, so that fine-tuning can be completely avoided for some pruning steps.

To implement this idea, ICE-Pruning makes three key changes to the standard pruning-fine-tuning pipeline.
First, instead of fine-tuning the full DNN model after every pruning step, ICE-Pruning measures the accuracy drop immediately after a pruning step; it then skips fine-tuning if the accuracy drop is less than a certain threshold.
Second, we introduce a novel layer freezing strategy to reduce fine-tuning overheads in each pruning step; this automatically determines which layers have minimal effect during fine-tuning, and can therefore keep their retained weights fixed during fine-tuning.
Third, we introduce a custom pruning-aware learning rate scheduler that dynamically adjusts the learning rate based on the current pruning level; this improves the accuracy after each fine-tuning stage and thus reduces the number of times fine-tuning is triggered overall.
Since these three changes introduce new hyperparameters, we also design an efficient auto-tuning stage with a custom objective function that simultaneously minimizes the accuracy loss and the time consumption, thus avoiding the need for manual tuning. 

Note that ICE-Pruning does not introduce any new pruning criteria. Instead, it supports widely-used existing criteria (e.g., L1 norm~\cite{l1}, random, entropy-based~\cite{entropy} and mean activation~\cite{molchanov2017pruning} filter pruning), thus providing a plug-and-play solution to improve pruning time on existing pipelines. 

The contributions of the paper include the following:

\begin{itemize} 
    \item ICE-Pruning, a novel pipeline for accelerating DNN pruning that is based on three key components: i) a mechanism to determine when fine-tuning should be performed based on an accuracy drop threshold; ii) a novel layer freezing strategy to reduce fine-tuning overheads in each pruning step; iii) a pruning-aware learning rate scheduler to further improve the accuracy of each pruning step and reduce the possibility of triggering fine-tuning.
    
    \item An efficient auto-tuning stage to avoid the need to manually tune pruning hyperparameters. 
   
    \item An evaluation of ICE-Pruning with different pruning criteria, demonstrating that it works with all of them and also identifying the criterion that yields the best results.

    \item An ablation study to show how the different components of the ICE-Pruning pipeline impact both accuracy and overall pruning time.

    \item A comparison of ICE-Pruning against i) a baseline iterative pruning pipeline and ii) the state-of-the-art method AAP, using three different combinations of DNN model and dataset. 
    Overall our results demonstrate that ICE-Pruning can accelerate pruning by up to $9.61\times$ while maintaining similar (and sometimes better) final accuracy than existing solutions. 
\end{itemize}

The rest of the paper is organized as follows: In Section~\ref{sec:related}, we provide the required background and discuss related works; Section~\ref{sec:method} shows the detailed design of ICE-Pruning; in Section~\ref{sec:exper}, we evaluate ICE-Pruning and provide an ablation study of its components and a comparison with the state-of-the-art; finally, Section~\ref{sec:conclusion} provides some conclusions and briefly discusses potential future work.

\section{Background and Related Work}
\label{sec:related}

%***************************************************************************************************

\subsection{Deep Neural Network pruning}

DNN pruning methods can be characterized with four key dimensions~\cite{blalockWhatStateNeural2020}: \emph{Structure}, \emph{Scoring}, \emph{Scheduling}, and \emph{Fine-tuning}.

\begin{itemize}
    \item \emph{Structure} refers to the level of granularity of pruning, and is generally divided into two classes: unstructured and structured.
    In unstructured pruning individual weights are removed, whereas structured pruning removes groups of parameters such as whole filters/channels or attention heads. With the increasing attention on pruning research~\cite{hoefler2021sparsity}, several approaches have been developed~\cite{blalockWhatStateNeural2020,deng2020model}. 
    Our ICE-Pruning pipeline focuses on structured pruning.

    \item \emph{Scoring} refers to which parameters to prune.
    For example, a popular approach is to use the L1 norm~\cite{l1}, where we prune parameters with absolute values closer to zero. The L2 norm is an alternative solution to determine which structures should be pruned~\cite{he2018soft}. The similarity of the structures within the same layers can also be used to make the pruning decisions~\cite{song2022filter}. For Transformers, previous research~\cite{michel2019sixteen} shows that attention heads can also be pruned guided by the Taylor Expansion~\cite{molchanov2016pruning} of the difference between before and after pruning the heads. 
    ICE-Pruning supports various scoring approaches such as L1 norm.

    \item \emph{Scheduling} refers to how much we prune in each step.
    For example, we might perform all the pruning in a single step, one-shot pruning~\cite{l1}; or gradually increase the amount of pruning in iterative steps, iterative pruning~\cite{LRF}. 
    ICE-Pruning is based on the iterative pruning.
    
    \item Finally, \emph{fine-tuning} refers to how we recover lost accuracy by training the model for a small number of epochs. The typical method is to use the training dataset to do the training~\cite{LRF}, and ICE-Pruning does this.
\end{itemize}

%***************************************************************************************************

\subsection{Fine-tuning}

As compared to one-shot pruning, iterative pruning can yield better accuracy if we fine-tune after every pruning step~\cite{l1}.
However, the larger the DNN model is, the more expensive fine-tuning becomes.
Therefore, many methods only fine-tune for 1 or 2 epochs with some accuracy penalty, sometimes training more epochs for the final pruning step~\cite{LRF,entropy,ThinNet}.
Note that even with a low number of epochs, the fine-tuning process can still be prohibitively time-consuming due to the large dataset sizes, the model sizes or the limited computing capacity available. 

To avoid the high cost of fine-tuning, ICE-Pick~\cite{hu2023icepick} tries to reduce the overall time consumption while other solutions entirely eliminate fine-tuning from the pruning process.
For example, NFP~\cite{liu2020nfp} compensates the pruned layers with their Batch Normalization beta factors instead of fine-tuning the model. 
Roy et al.~\cite{roy2020pruning} proposes moving pruning to training to avoid fine-tuning during pruning. 
However, such solutions require specialized training operations (e.g., NFP uses a customized loss function), which has limited applicability to practical DNN model deployment where users already have a pre-trained model (common in modern artificial intelligence tasks such as natural language processing and computer vision). %Furthermore, since the pre-training paradigm is dominating in modern artificial intelligence tasks such as natural language processing and computer vision, this limitation is more severe.

It is important to note that fine-tuning is not exclusive to pruning. 
In transfer learning~\cite{brock2017freezeout}, fine-tuning is used to adapt models to domains with limited data. Similar to pruning, fine-tuning in transfer learning can also be very time-consuming~\cite{liu2021autofreeze}.
To address this problem, some methods use layer freezing to accelerate the fine-tuning process~\cite{brock2017freezeout,lee2019would,liu2021autofreeze,topcuoglu2023local}, where the parameters of some layers of the model are `frozen' such that they cannot be changed during fine-tuning. 
We also apply this idea to model pruning, due to the similar role of fine-tuning in both tasks. Note also that in both tasks, fine-tuning tries to adapt the models into new states; for transfer learning, fine-tuning adapts the models to new datasets; for pruning, fine-tuning adapts the models to new sparsity structures.

%***************************************************************************************************

\begin{figure*}[t]
\centering
\begin{subfigure}{0.90\linewidth}
    \includegraphics[width=1\textwidth]{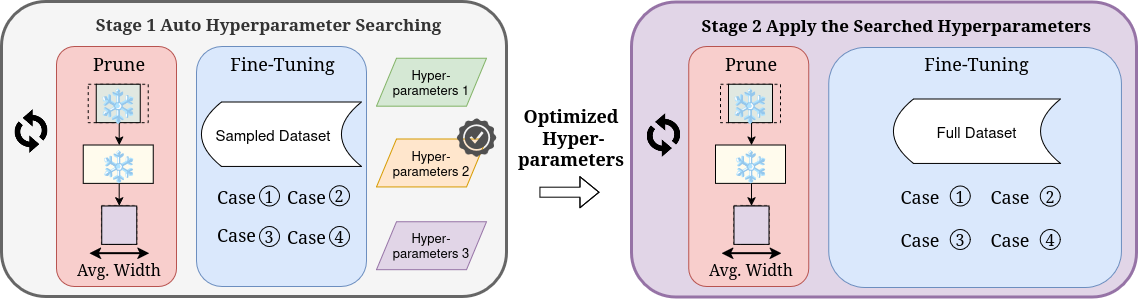}
       \vspace{-0.6cm}
       \caption{Overview}
       \vspace{0.2cm}
      \label{fig:method-overview}
\end{subfigure}
\begin{subfigure}{0.90\linewidth}
    \includegraphics[width=1\textwidth]{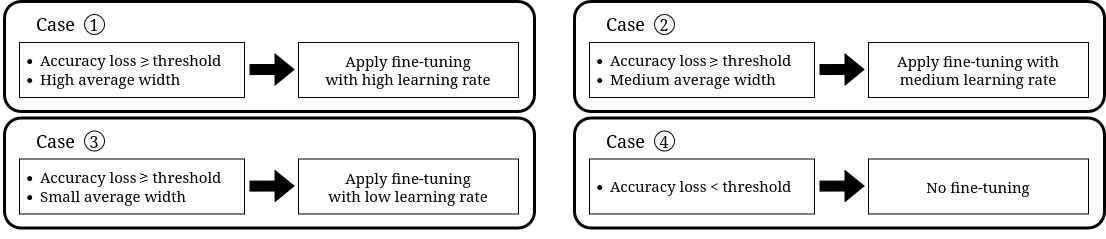}
       \vspace{-0.7cm}
       \caption{Fine-Tuning Cases}
      \label{fig:fine-tuning}
\end{subfigure}
\caption{ICE-Pruning pipeline: (a) Using subsampled datasets, we automatically search for the best hyperparameters by running the whole pruning pipeline with the current hyperparameters (Stage 1). Then, we re-run the pruning pipeline with the full dataset and the best hyperparameters found (Stage 2). In each step, we prune the model (note that the dotted lines are the edges of the pruned parts) and then we move to the next pruning step or fine-tune the model by applying the freezing strategy; (b) Fine-tuning has four cases depending on the accuracy loss and pruned model width. The learning rate is adjusted dynamically, and fine-tuning for a given step is halted if the accuracy loss is lower than the threshold.}
\label{fig:method}
\end{figure*}

\subsection{Automatic pruning pipeline}

Some pruning pipelines such as AAP~\cite{AAP} aim to automatically prune DNN models~\cite{AAP}, but do not focus on efficiency. 
AAP is based on LTH~\cite{frankle2018lottery}, which shows a model that includes sub-models that are easier to train. As a result, AAP tries to find the best state (the ideal sub-model for training) of the DNN models to prune and fine-tune. 
To do that, AAP rewinds the weights of the pruned model to previous states if the current state cannot satisfy, for example, the target accuracy. This makes the running time even longer. Also, for models that have high parameter redundancies, rewinding to the previous state is not necessarily better than direct fine-tuning on the current state, as the high number of remaining parameters can still have enough expressivity~\cite{pmlr-v139-liu21y}.

\section{ICE-Pruning}
\label{sec:method}

\subsection{Overview}

The main idea behind ICE-Pruning is that instead of fine-tuning a full DNN model in every pruning step, we freeze less sensitive layers and skip fine-tuning when the accuracy loss is lower than a user-defined threshold.
Figure~\ref{fig:method} gives an overview of our ICE-Pruning pipeline, with Algorithm~\ref{alg:FIR} describing the steps in more detail.

As we can see in Figure~\ref{fig:method-overview}, ICE-Pruning has two main stages.
In Stage 1, we automatically search for the best hyperparameters using subsampled datasets. Using the current candidate hyperparameters, we iteratively prune the DNN model. After applying each pruning step, we move to the next pruning step or fine-tune the non-frozen layers of the model by applying the freezing strategy. If fine-tuning happens, we adjust the learning rate dynamically. 
In Stage 2, we redo the iterative pruning of the previous stage but using the full dataset and applying the best hyperparameters found. 
During both stages of fine-tuning (Figure~\ref{fig:fine-tuning}), cases \Circled{1}, \Circled{2}, and \Circled{3} show how we gradually reduce the learning rate as the level of pruning increases. In case \Circled{4} we stop fine-tuning when the accuracy loss is below a threshold. 

In Algorithm~\ref{alg:FIR}, Stage 1 (lines 4-12) iterates the candidate hyperparameters and selects the ones that minimize the errors (line 9), which involves both accuracy loss and time consumption during the pruning-fine-tuning (PFT) operations. This process (lines 6-12) corresponds to Equation~\ref{eqn:optf}. 
Then in Stage 2 (line 14), the pruning-fine-tuning is applied again but with the best hyperparameters found. 
Note that PFT is defined in Algorithm~\ref{alg:FIr_ops}, where the model is first frozen (line 5) by our novel freezing strategy (discussed in Section~\ref{sec:method:freezing}). 
Then, for each pruning step, the algorithm measures the accuracy drop value (line 8). 
If the value is larger or equal than the pre-defined threshold (line 8), it fine-tunes the model by adjusting the learning rate (lines 9-11). The details of lines 6-11 will be discussed in Section~\ref{sec:method:PFT}. 
The following subsections discuss the key design aspects of the algorithm and further motivate our design choices.

%***************************************************************************************************

\subsection{Layer Freezing}
\label{sec:method:freezing}

We adopt \textit{layer freezing}, a class of techniques commonly used in transfer learning~\cite{brock2017freezeout,lee2019would,liu2021autofreeze}.
This is motivated by the fact that some DNN layers (typically initial layers in the model) see smaller weight shifts than others during fine-tuning, and the ranks of the weight shifts are generally maintained (we provide evidence of this in Section~\ref{subsec:freezing_results}). 
Therefore, we can skip training these layers and reduce the fine-tuning time. 
Also, since the ranks of the shifts are generally maintained, we can get information of which layers are less sensitive to pruning from a single pruning step. 
Based on those insights, we design the following \emph{freezing strategy}: We run a pruning step followed by fine-tuning while tracking the weight changes of different layers. After fine-tuning, and based on the input freezing percentage (i.e., the proportion of frozen layers), the layers with smaller weight changes are frozen. 
To eliminate the impact of weight magnitudes and have fair comparisons, weight changes of each layer are normalized by the norm of the layer weights. This is a novel freezing strategy in our work, and we conduct an ablation study to show the effectiveness of this component in Section~\ref{subsec:ablations_study}. 
In Figure~\ref{fig:method}, we use a snowflake to represent the layer freezing. %To the best of our knowledge, layer freezing have not been applied in the model pruning domain before. 

\newcommand{\Select}[1]{\State \textbf{select} #1}
\newcommand{\Stage}[1]{\State \textbf{Stage} \textbf{#1}}

\begin{algorithm}[t]
\begin{algorithmic}[1]
%\footnotesize
\State {\textbf{Input:} 

           (1)~Pruning-Fine-Tuning operator $\mathrm{PFT}(...)$,
           
           (2)~Pruning steps $\mathrm{\sigma}$,
           
           (3)~Model $\mathrm{M}$,
           
           (4)~Hyperparameters Search Space $\mathrm{\Lambda}$,

           (5)~Accuracy testing operator $\mathrm{Test}(.,.)$,

           (6)~DataSet $\mathrm{D}$,

           (7)~DataSet sampler $\mathrm{Sp}(.)$,

           (8)~Layer-wise Pruning operator $\mathrm{P}(.)$,
           }
           
\State {\bfseries Output:} Pruned Model $\mathrm{M^*}$\;
\Stage{1: Auto-Tuning the Hyperparameters}
\State $\mathrm{acc}_{\mathrm{orig}} \leftarrow \mathrm{Test}(\mathrm{M}, \mathrm{Sp}(\mathrm{D}))$\;
\State $\mathrm{e}_\mathrm{best} \leftarrow \mathrm{MAX\_VALUE}$\;
\For{$\lambda \in \Lambda$}
      \State $\mathrm{time}, \mathrm{acc} \leftarrow \mathrm{PFT}(\mathrm{\sigma}, \mathrm{M}, \mathrm{\lambda}, \mathrm{Sp}(\mathrm{D}), \mathrm{P})$\;
      \State $\mathrm{\Delta_{A}} \leftarrow \mathrm{acc}_\mathrm{orig} - \mathrm{acc}$\;
      \Comment{eq.~\ref{eqn:delta}~~~~~}
      \State $\mathrm{e} \leftarrow \frac{\mathrm{time} + \mathrm{\Delta_{A}}}{\max(\mathrm{time}, \mathrm{\Delta_{A}})}$\;
      \If{$\mathrm{e} < \mathrm{e}_\mathrm{best}$}
        \State $\mathrm{\lambda_{best}} \leftarrow \mathrm{\lambda}$\;
        \State $\mathrm{e_{best}} \leftarrow \mathrm{e}$\;
      \EndIf
\EndFor
 
\Stage{2: Apply the Searched Hyperparameters}
\State $\mathrm{M^*} \leftarrow \mathrm{PFT}(\mathrm{\sigma}, \mathrm{M}, \mathrm{\lambda_{best}}, \mathrm{D}, \mathrm{P})$;
\State \textbf{return} $\mathrm{M^*}$\;
\caption{ICE-Pruning Pipeline}
\label{alg:FIR}
\end{algorithmic}
\end{algorithm}
% Define custom indentation command
\newcommand{\myindent}{\hspace*{\algorithmicindent}}

% Define custom command for 'with' clause in Python
\newcommand{\With}[2]{%
    \State \textbf{with} 
        #1 
    \textbf{do} \\%
    \myindent%
    \begin{minipage}[t]{\dimexpr\linewidth-\algorithmicindent}
        #2 
    \end{minipage} 
    \\
    %\textbf{end with}
}

\begin{algorithm}[t]
\begin{algorithmic}[1]
%\footnotesize
\State {\textbf{Input:} 

          (1)~Layer-wise Pruning operator $\mathrm{P}(.)$,

           (2)~Pruning steps $\mathrm{\sigma}$,
           
           (3)~Freezing operator $\mathrm{Freeze}(.)$,
           
           %(2)~Freeze operator $\mathrm{Freeze}(.)$,
           (4) Accuracy testing operator $\mathrm{Test}(.,.)$,
           
           (5)~Parameter calculating operator $\mathrm{Pr}(.)$, 
           
           (6) Fine-tuning operator $\mathrm{FT}(.,.)$, 
           
           (7)~Model $\mathrm{M}$, 

           (8)~Dataset $\mathrm{D}$,
           
           (9)~Original accuracy $\mathrm{acc_{orig}}$, 
           
           (10)~Searched Hyperparameters $\mathrm{\lambda}$
           
            ~~(10.1)~Accuracy drop threshold $\mathrm{\theta}$, 
           
            ~~(10.2)~Frozen layers percentage $\mathrm{\eta}$, 
           
            ~~(10.3)~Initial learning rate (LR) $\mathrm{LR_{base}}$, 
           
            ~~(10.4)~Maximum LR change $\mathrm{\Delta}$, 
           
            ~~(10.5)~Pruning level $\mathrm{p}$ that $\mathrm{\Delta}$ midpoint is reached, 
           
            ~~(10.6)~The shape control $\mathrm{\beta}$}
          \State {\bfseries Output:} Pruned Model $\mathrm{M^*}$\;

\State $\mathrm{M_{orig}} \leftarrow \mathrm{M}$\;
\State $\mathrm{acc_{orig}} \leftarrow \mathrm{Test}(\mathrm{M}, \mathrm{D})$\;
 %\With{the context of applying $F$}{
 %\State Apply the $F$;
 \State $\mathrm{M_{frozen}} \leftarrow \mathrm{Freeze}(\mathrm{M}, \mathrm{\eta})$\;
 \For{$step \in \{1\,\ldots \mathrm{\sigma}\} $} 
    \State $\mathrm{M_{frozen}} \leftarrow \text{P}(\mathrm{M_{frozen}}, \mathrm{step})$\;
    \If{$\mathrm{acc_{orig}} - \mathrm{Test}(\mathrm{M_{frozen}}, \mathrm{D}) \geqslant \mathrm{\theta}$} 
    \State $\mathrm{\alpha} \leftarrow \frac{\mathrm{Pr}(\mathrm{M_{frozen}})}{\mathrm{Pr}(\mathrm{M_{orig}})}$ \;
       \State $\mathrm{LR_{max}} \leftarrow \mathrm{LR_{base}} - \frac{\mathrm{\Delta}}{1 + \left(\frac{\mathrm{\alpha}}{2 \times (1 - \mathrm{p}) - \mathrm{\alpha})}\right)^{\mathrm{\beta}}}$ \Comment{eq.~\ref{eqn:max_lr}~~~~~}
       %\mathrm{Eq.~\ref{eqn:max_lr}}(\Delta,\alpha,p,\mathrm{base})$ 
       \State $\mathrm{FT}(\mathrm{M_{frozen}}, \mathrm{LR_{max}}, \mathrm{D}) $\;
       \EndIf
       \EndFor
    
    \State $\mathrm{M^*} \leftarrow \mathrm{M_{frozen}}$\;
    \State \textbf{return} $\mathrm{M^*}$\;
%    }
\caption{Pruning-Fine-Tuning Operator (PFT)}
\label{alg:FIr_ops}
\end{algorithmic}
\end{algorithm}

%***************************************************************************************************

\subsection{Pruning and Fine-tuning}
\label{sec:method:PFT}

The main goal of ICE-Pruning is to iteratively prune DNN models but to trigger fine-tuning as few times as possible, thus reducing the overall pruning time.
Lines 7-11 of Algorithm~\ref{alg:FIr_ops} show a pruning and fine-tuning step.
For each layer in the model, including the frozen ones, we apply structured pruning and then we evaluate the model on the test dataset. 
If we observe an accuracy drop higher than or equal to our threshold (line 8 of Algorithm~\ref{alg:FIr_ops}), then we trigger fine-tuning; otherwise, we skip it.
This accuracy threshold is searched automatically or set manually, with its value varying depending on the learning task and the application's tolerance to accuracy loss. 

Note that, since our method acts as a meta-algorithm that wraps existing pruning criteria, we do not design a new pruning criterion. However, to select the pruning criterion that works best with ICE-Pruning, we compare four popular criteria in Section~\ref{subsec:compare_pc}. 

If we trigger fine-tuning, we want to minimize the training time while converging to a higher accuracy more quickly.
Dynamic learning rates are widely used to adapt to changing learning conditions~\cite{li2019towards,liu2019variance}.
It has been shown that narrower models (i.e., smaller average width, where we define `width' as the number of non-pruned structures in a given layer) have a narrower loss landscape~\cite{vis}, and thus can benefit from lower maximum learning rates. 
When we prune, we similarly shrink the loss landscape. That means that we benefit from controlling the upper bound of whatever learning rate scheduler is being used for fine-tuning, to match the current loss landscape.
%Furthermore, high learning rates can lead to over-fitting~\cite{smith2018disciplined}, which is an example of mismatch between accuracy loss and learning rate~\cite{sun2021learning}.
Therefore, inspired by S-Cyc~\cite{s-cyc}, we design a pruning-aware scheduler to choose the \textit{maximum} learning rate, based on the current level of pruning, measured according to the average width. Our maximum learning rate is defined by:
\begin{equation}
       \mathrm{LR}_\mathrm{max} = \mathrm{LR}_\mathrm{base} - \frac{\mathrm{\Delta}}{1 + \left(\frac{\alpha}{2 \times (1 - \mathrm{p}) - \mathrm{\alpha})}\right)^{\mathrm{\beta}}} 
\label{eqn:max_lr}
\end{equation}
where $\mathrm{LR}_\mathrm{max}$ is the upper bound of the existing scheduler used for fine-tuning (e.g., cosine decay or constant), and $\alpha$ is the proportion of unpruned parameters. 
The hyperparameters are:        
$\mathrm{LR}_\mathit{base}$, the maximum learning rate to be used for the unpruned model;
$\Delta$, the maximum change in the learning rate during pruning, with range (0, $\mathrm{LR}_\mathit{base}$); 
$p$, the level of pruning where the midpoint of $\Delta$ is reached, with range (0, 50\%]; 
and $\beta$, which controls the shape of the curve, with range (0, $+\infty$). 
By tuning these hyperparameters, we can flexibly formulate different types of decreasing learning schedules for different situations. We conduct an ablation study to show the impact of this component in Section~\ref{subsec:ablations_study}.

%***************************************************************************************************

\subsection{Automatic Selection of hyperparameters}

The behavior of ICE-Pruning is determined by several hyperparameters. To avoid the need for manually tuning these hyperparameters, we add an auto-tuning stage. We empirically design the following optimization cost function:
\begin{equation}
      \mathrm{\lambda} = \mathop{\arg\min}\limits_{\mathrm{\lambda}} \frac{\mathrm{PT} + \mathrm{\Delta_{A}}}{\max(\mathrm{PT}, \mathrm{\Delta_{A})}}
\label{eqn:optf}
\end{equation}
%\mathit{Acc}_\mathit{original} - \mathit{Acc}_\mathit{pruned}
%
where
\begin{equation}
    \mathrm{\Delta_{A}} = \mathrm{Acc_{original}} - \mathrm{Acc_{pruned}}
\label{eqn:delta}
\end{equation}
and $\mathrm{\lambda}$ is the set of hyperparameters; 
$PT$ is the pruning time, which is the overall time (in seconds) required by the whole pruning-fine-tuning from the first layer to the last; 
$\mathrm{Acc_{original}}$ is the accuracy of the full model; 
and $\mathrm{Acc_{pruned}}$ is the final accuracy of the pruned model. 
We normalize the sum of $\mathrm{PT}$ and $\mathrm{\Delta_{A}}$ by the larger value of them to avoid the domination of larger values.
Empirically, since accuracy $\mathrm{PT}$ is always larger than $\mathrm{\Delta_{A}}$, minimizing this function is equivalent to minimizing the accuracy loss per time-unit (e.g., seconds) during pruning.
%can ensure that the searched hyperparameters serve the goal of ICE-Pruning, i.e., using the minimum time to prune DNN models while getting as high as possible final accuracy. 
We use a search algorithm to find the optimal $\mathrm{\lambda}$ instead of gradient-based methods. In this way, the pruning time does not need to be differentiable, which is required by gradient-based methods.

We compared several popular hyperparameter search algorithms (i.e., Grid Search, Tree-structured Parzen Estimator~\cite{TPE}, Random Sampling, Gaussian process-based Bayesian Optimization~\cite{optuna_2019}) and found that grid search performs best, since our search space is low-dimensional.
% it does not have combination explosion if the searching trails and searching space are small.
To save time in the auto-tuning stage, we only use a randomly sampled subset of the data. The sampled data has a similar distribution as the whole dataset, thus if the hyperparameters optimize the objective (\ref{eqn:optf}) with respect to the sampled dataset, they should also optimize it with respect to the full dataset, assuming the same DNN models and pruning methods. 
As a result, the optimal hyperparameters can be transferred to the later stage. Our experimental results in Section\ref{sec:exper} show that this design works.

\section{Evaluation}
\label{sec:exper}

In this section, we validate ICE-Pruning and demonstrate its ability to reduce the overall pruning time while maintaining accuracy.
First, we show how some DNN layers change much less than others during fine-tuning, justifying the use of layer freezing (Section~\ref{subsec:freezing_results}). 
We then compare ICE-Pruning when using four different pruning criteria (L1 norm filter pruning, Random filter pruning, Entropy-based filter pruning and Mean activation filter pruning) and justify which one works best with ICE-Pruning (Section~\ref{subsec:compare_pc}).
After that, we conduct an ablation study to explore the impact of different components such as the accuracy drop threshold mechanism, layer freezing, and the pruning-aware learning rate scheduler (Section~\ref{subsec:ablations_study}).
Finally, we compare ICE-Pruning with a baseline iterative pruning pipeline and the state-of-the-art pipeline AAP~\cite{AAP} for different datasets and DNN models (Section~\ref{subsec:sotas}).

%***************************************************************************************************

\subsection{Experimental setup}

We conduct experiments with the following datasets: i) CIFAR-10~\cite{cifar-10} (using the full dataset when doing the fine-tuning, except for the hyperparameter search phase) and ii) TinyImageNet~\cite{le2015tiny} and ImageNet~\cite{IM} (using subsampled datasets when performing the fine-tuning). 
Note that for all the experiments we adopt the common practice of performing only one fine-tuning epoch in each pruning step~\cite{LRF,entropy,ThinNet}, except the non-AAP experiments for TinyImageNet, in which after fine-tuning in the last pruning step, we applied four more epochs with full datasets.
For CIFAR-10, all pipelines use knowledge distillation~\cite{kd,turnerDistilling2018}, following the practice of Joo et al.~\cite{LRF}.

We use L1 norm filter pruning~\cite{l1} in each non-AAP experiment (including Section~\ref{sec:method:freezing}), except Section~\ref{subsec:compare_pc} where we compare different heuristics. 
All the models are pruned layer by layer with consistent specified pruning ratios for each layer in each experiment. The pruned layers in our DNN models are the basic blocks in ResNets~\cite{resNet}, the dense blocks in DenseNet~\cite{huang2017densely}, and the bottleneck block in Wide ResNet~(WRN)~\cite{zagoruyko2017wideresidualnetworks}. 
For fine-tuning, all experiments use the SGD~\cite{ruder2016overview} optimizer, with batch size of 128 or 256, base learning rate of 0.001 (except the experiments that need to search for it), momentum of 0.9 and weight decay of 0.0001. 
The details of the fine-tuning are shown in Table~\ref{tab:exp_set}.
%The other details of the setting ups are in the Table~\ref{tab:exp_set}.

\begin{table}[t]
\begin{center}
\caption{Fine-tuning details for different model and dataset. BS is batch size, LR is base learning rate, MO is Momentum, WD is weight decay. Note that the TinyImageNet experiments use batch size 256 for AAP and 128 for the rest pipelines.}
\begin{tabular}{cccccc}
\toprule
\textbf{Model} & \textbf{Dataset} & \textbf{BS} & \textbf{LR} & \textbf{MO} & \textbf{WD}\\ 
\midrule
ResNet-152 & CIFAR-10 & 128 &  0.001 & 0.9 & 0.0001 \\
DenseNet-121 & TinyImageNet & 128/256 &  0.001 & 0.9 & 0.0001\\
WRN-101 & ImageNet & 256 &  0.001 & 0.9 & 0.0001\\
\bottomrule
%\hline
\end{tabular}
\label{tab:exp_set}
\end{center}
\end{table}

We use PyTorch~\cite{paszke2019pytorch} as our main machine learning framework, with Optuna~\cite{optuna_2019} for automated hyperparameter search. The experiments were run on a single NVIDIA GeForce RTX 3090 GPU, except for ImageNet which run on two of them.

%***************************************************************************************************

\subsection{Justification of Layer Freezing}
\label{subsec:freezing_results}

\begin{figure}[t]
  \centering
  \begin{subfigure}{0.85\linewidth}
    %\fbox{\rule{0pt}{0.5in} \rule{.9\linewidth}{0pt}}
    %\includegraphics[width=1\linewidth]{fig/extension_figs/exp1_152_plot.png}
    \includegraphics[width=1\linewidth]{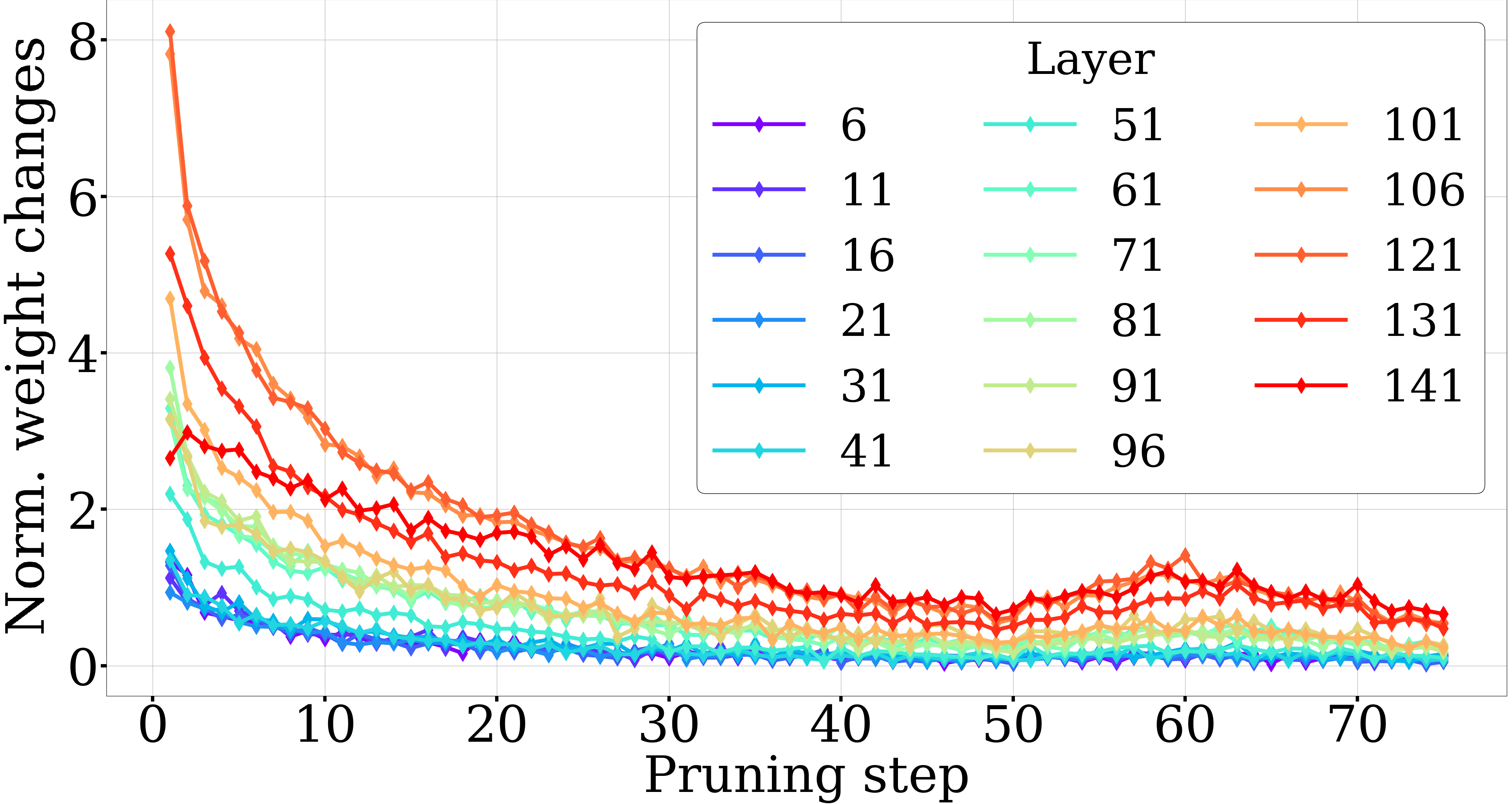}
    \caption{ResNet-152}
  \end{subfigure}
  \hfill
  \begin{subfigure}{0.85\linewidth}
  \vspace{6pt}
    %\fbox{\rule{0pt}{0.5in} \rule{.9\linewidth}{0pt}}
    \includegraphics[width=1\linewidth]{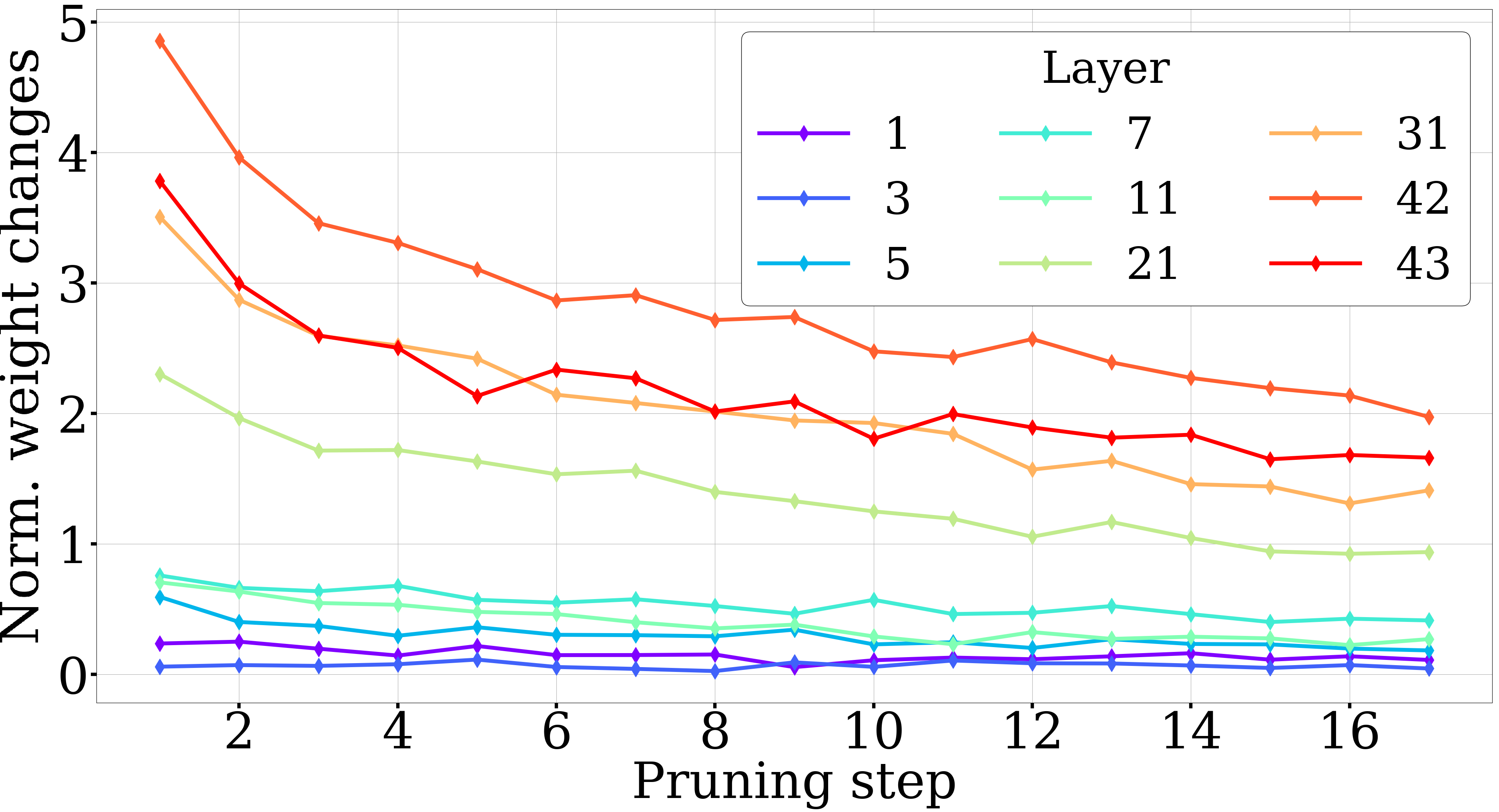}
    \caption{MobileNetV2}
  \end{subfigure}  
  \caption{Normalized weight changes ($\mathrm{\Delta W }\;/\; \|\mathrm{W_0}\|_2$) per layer for each pruning step in DNN iterative pruning; $\mathrm{\Delta W}$ is the weight changes, $\|\mathrm{W_0}\|_2$ is the L2 norm of the initial weights. The learning rate is 0.001 and the pruning ratio is 30\%.}
  \label{fig:shifting}
\end{figure}

We first study how the L1 norm of model weights changes during iterative fine-tuning, to justify the novel layer freezing strategies introduced in Section~\ref{sec:method:freezing}.
Specifically, we consider ResNet-152~\cite{resNet} and MobileNetV2~\cite{howardMobileNetsEfficientConvolutional2017} trained on CIFAR-10~\cite{cifar-10}. 
To eliminate the possibility that larger magnitude weights may have higher L1 norm changes, we normalize the results by the L2 norms of initial layer weights. Each experiment applies a pruning ratio (the proportion of filters removed) of 30\%. 
Figure~\ref{fig:shifting} shows how the weights of different layers change as the amount of fine-tuning increases.
We observe that across our DNN models, some layers tend to see smaller changes in weights during fine-tuning. This implies that those layers are insensitive to fine-tuning, or at least it is not worth spending time on them for fine-tuning because of the small changes.
We also observe that the ranks of the weight changes of the layers are generally maintained, so the sensitivity ranking of layers can be determined after the very first pruning step. 
This is a powerful insight, since determining which layers are less sensitive based on only one pruning step is cheap, and hence makes ICE-Pruning faster. 
In summary, this experiment justifies both freezing the less sensitive layers (determined by observing weight changes from one pruning step) and applying the freezing step only once.

%***************************************************************************************************

\subsection{Comparisons of different pruning criteria}
\label{subsec:compare_pc}

To know which pruning criterion works best with ICE-Pruning, we now compare four different pruning criteria:

\begin{itemize}
    \item \textbf{L1 norm filter pruning}: For each layer, the filters that have a smaller L1 norm will be pruned.

    \item \textbf{Random filter pruning}: For each layer, the filters are randomly pruned.

    \item \textbf{Entropy-based filter pruning~\cite{entropy}}: For each layer, the filters that correspond to lower feature map entropy values are removed.

    \item \textbf{Mean activation filter pruning~\cite{molchanov2017pruning}}: For each layer, the filters that correspond to the lowest mean activation values are removed.
\end{itemize}

For each pruning criterion, we conducted an experiment with ResNet-152 on CIFAR-10 and DenseNet-121~\cite{huang2017densely} on TinyImageNet~\cite{le2015tiny}. Each experiment applies pruning ratios of 60\%, 70\% and 80\%.

\begin{figure}[t]
  \centering
  \includegraphics[width=1\linewidth]{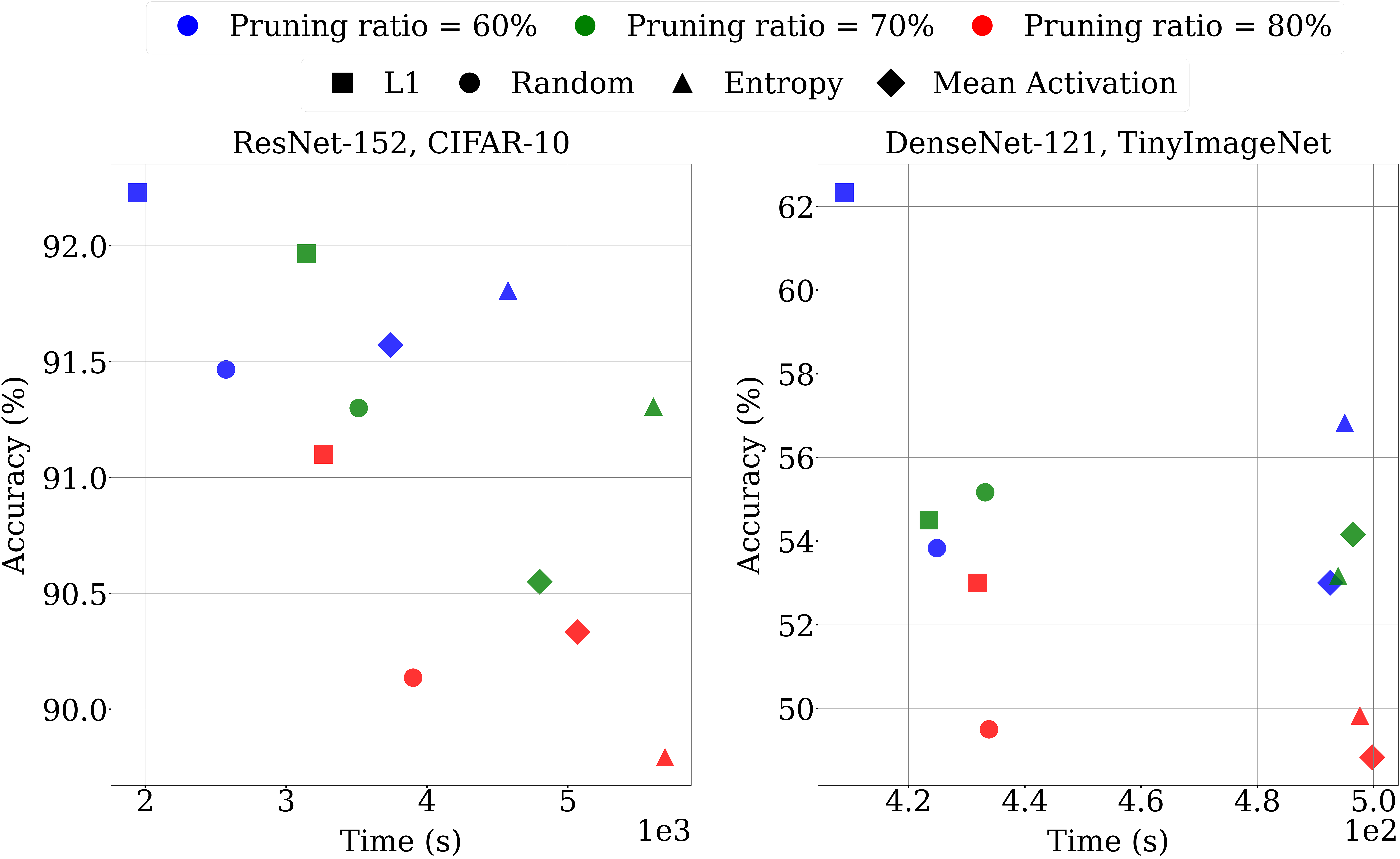}
  \caption{Pruning time vs. accuracy of ICE-Pruning with four different pruning criteria for two different models, Resnet-152 on CIFAR-10, and DenseNet-121 on TinyImageNet.}
  \label{fig:compare_p}
\end{figure}

In Figure~\ref{fig:compare_p}, we observe that for different DNN models, datasets and pruning ratios, L1 norm filter pruning provides the best trade-off between accuracy and time consumption. 
As a result, we use L1 norm as the pruning criterion for ICE-Pruning in the remaining experiments. Interestingly, we observe that random pruning performs well too. This is because of the low overhead of using this pruning criterion, and because the models have enough redundancies for these datasets. Therefore, we suggest users to use random pruning as well.

%***************************************************************************************************

\subsection{Ablation study of key components}
\label{subsec:ablations_study}

We explore the impact of the three key components of ICE-Pruning (accuracy drop threshold, freezing percentage, and learning rate scheduler) in order to provide deeper insights into our method. 
We select ResNet-152 on CIFAR-10 and DenseNet-121 on TinyImageNet. For each model, we apply ICE-Pruning but without one component and compare the accuracy and time consumption with the full pipeline. The pruning ratio is 60\%.

\begin{table}[h!]
\caption{Ablation study of ICE-Pruning key components. Threshold is the accuracy drop threshold mechanism, Freezing is layer freezing and Scheduler is pruning-aware scheduler. Time is the time consumption.}
\begin{subtable}{0.5\textwidth}
  %\begin{center}
    \caption{Ablation study with Resnet-152 on CIFAR-10}
    \label{tab:abl:table1}
    \begin{tabular}
     {p{0.15\linewidth}|p{0.15\linewidth}|p{0.15\linewidth}|p{0.17\linewidth}|p{0.13\linewidth}}
     \toprule
      \textbf{Threshold} & \textbf{Freezing} & \textbf{Scheduler} & \textbf{Accuracy(\%)} & \textbf{Time(s)}\\ 
      \midrule
      \checkmark & \checkmark & \checkmark &  92.23 & 1943 \\
       \ding{55} & \checkmark & \checkmark &  92.81 & 4459 \\
      \checkmark & \ding{55} & \checkmark &  92.30 & 2177 \\
      \checkmark & \checkmark & \ding{55} &  91.42 & 2200 \\
      \bottomrule
    \end{tabular}
  %\end{center}
\end{subtable}
\hfill
\begin{subtable}{0.5\textwidth}
  %\begin{center}
\vspace{10pt}
    \caption{Ablation study with DenseNet-121 on TinyImageNet}
    \label{tab:abl:table2}
    \begin{tabular}
     {p{0.15\linewidth}|p{0.15\linewidth}|p{0.15\linewidth}|p{0.17\linewidth}|p{0.13\linewidth}}
     \toprule
      \textbf{Threshold} & \textbf{Freezing} & \textbf{Scheduler} & \textbf{Accuracy(\%)} & \textbf{Time(s)}\\ 
      \midrule
      \checkmark & \checkmark & \checkmark &  62.33 & 409 \\
       \ding{55} & \checkmark & \checkmark &  62.17 & 426 \\
      \checkmark & \ding{55} & \checkmark &  58.67 & 429 \\
      \checkmark & \checkmark & \ding{55} & 61.83 & 408 \\
      \bottomrule
    \end{tabular}
  %\end{center}
\end{subtable}
\label{tab:abl}
\end{table}

From Table~\ref{tab:abl}, we observe that for both DNN models, using all ICE-Pruning components provides the best accuracy and pruning time trade-off. 
We also observe that the mechanism based on accuracy drop threshold provides the greatest time reduction. Specifically, the accuracy drop threshold mechanism provides up to 56.43\% time reduction, layer freezing provides up to 10.75\% time reduction and pruning-aware learning rate scheduler provides up to 11.68\% time reduction.

%***************************************************************************************************

\subsection{Comparison against existing pruning pipelines}
\label{subsec:sotas}

\begin{table}[t]
\caption{Comparisons of ICE-Pruning with AAP and Baseline. Acc is accuracy, Time is time consumption.}
\begin{subtable}{0.5\textwidth}
  %\begin{center}
    \caption{60\% pruning accuracy and time consumption }
    \label{tab:sota:table1}
    \begin{tabular}
     {p{0.16\linewidth}|p{0.18\linewidth}|p{0.17\linewidth}|p{0.1\linewidth}|p{0.13\linewidth}}
     \toprule
      \textbf{Pipeline} & \textbf{Model} & \textbf{Dataset} & \textbf{Acc(\%)} & \textbf{Time(s)}\\ 
      \midrule
      Baseline & ResNet-152 & CIFAR-10 & 92.15  & 3793 \\
      AAP & ResNet-152 & CIFAR-10 &  91.91 & 4909 \\
      ICE-Pruning & ResNet-152 & CIFAR-10 & 92.23  & \textbf{1943} \\
      \midrule
      Baseline & DenseNet-121 & TinyImageNet & 59.50  & 1212 \\
      AAP & DenseNet-121 & TinyImageNet & 54.53  & 898 \\
      ICE-Pruning & DenseNet-121 & TinyImageNet & 62.33  & \textbf{409} \\
      \midrule
      Baseline & WRN-101-2 & ImageNet & 62.77  & 10105 \\
      AAP & WRN-101-2 & ImageNet & 59.3  & 6935 \\
      ICE-Pruning & WRN-101-2 & ImageNet & 61.81  & \textbf{6530} \\

      \bottomrule
    \end{tabular}
  %\end{center}
\end{subtable}
\hfill
\begin{subtable}{0.5\textwidth}
  %\begin{center}
  \vspace{10pt}
    \caption{70\% pruning accuracy and time consumption }
    \label{tab:sota:table2}
    \begin{tabular}
     {p{0.16\linewidth}|p{0.18\linewidth}|p{0.17\linewidth}|p{0.1\linewidth}|p{0.13\linewidth}}
     \toprule
      \textbf{Pipeline} & \textbf{Model} & \textbf{Dataset} & \textbf{Acc(\%)} & \textbf{Time(s)}\\ 
      \midrule
      Baseline & ResNet-152 & CIFAR-10 & 91.29  & 3829 \\
      AAP & ResNet-152 & CIFAR-10 & 91.63  & 5301 \\
      ICE-Pruning & ResNet-152 & CIFAR-10 & 91.97  &  \textbf{3144}\\
      \midrule
      Baseline & DenseNet-121 & TinyImageNet &  57.00 & 1237 \\
      AAP & DenseNet-121 & TinyImageNet & 53.15  & 1744 \\
      ICE-Pruning & DenseNet-121 & TinyImageNet & 54.50  &  \textbf{423}\\
      \midrule
      Baseline & WRN-101-2 & ImageNet & 55.65 & 9764\\
      AAP & WRN-101-2 & ImageNet & 55.05  & 15250 \\
      ICE-Pruning & WRN-101-2 & ImageNet & 54.97  & \textbf{7393} \\
      
      \bottomrule
    \end{tabular}
  %\end{center}
\end{subtable}

\hfill
\begin{subtable}{0.5\textwidth}
  %\begin{center}
  \vspace{10pt}
    \caption{80\% pruning accuracy and time consumption }
    \label{tab:sota:table3}
    \begin{tabular}
     {p{0.16\linewidth}|p{0.18\linewidth}|p{0.17\linewidth}|p{0.1\linewidth}|p{0.13\linewidth}}
     \toprule
      \textbf{Pipeline} & \textbf{Model} & \textbf{Dataset} & \textbf{Acc(\%)} & \textbf{Time(s)}\\ 
      \midrule
      Baseline & ResNet-152 & CIFAR-10 & 91.11  & 3811 \\
      AAP & ResNet-152 & CIFAR-10 &  91.20 & 8719 \\
      ICE-Pruning & ResNet-152 & CIFAR-10 &  91.10 &  \textbf{3266}\\
      \midrule
      Baseline & DenseNet-121 & TinyImageNet &  50.50 & 1209 \\
      AAP & DenseNet-121 & TinyImageNet & 49.84  & 4151 \\
      ICE-Pruning & DenseNet-121 & TinyImageNet & 53.00 &  \textbf{432}\\
      \midrule
      Baseline & WRN-101-2 & ImageNet & 45.62 & 9344 \\
      AAP & WRN-101-2 & ImageNet & 44.01  & 14979 \\
      ICE-Pruning & WRN-101-2 & ImageNet & 46.85  & \textbf{8897} \\
      \bottomrule
    \end{tabular}
  %\end{center}
\end{subtable}
\label{tab:sota}
\end{table}

We now compare ICE-Pruning with a baseline and the state-of-the-art method AAP~\cite{AAP} in terms of final accuracy and time consumption. 
The baseline is the naive way of iterative pruning in which DNN models are pruned component by component, and at each pruning step an epoch of fine-tuning is applied~\cite{LRF,l1,entropy}. 
We conduct the experiments with three pruning ratios (60\%, 70\% and 80\%), with three DNN models (ResNet-152, DenseNet-121, and Wide ResNet-101-2 [WRN-101-2]) and three datasets (CIFAR-10, TinyImageNet and ImageNet). 
Note that AAP prunes the models globally, which means that it considers all the model components that are to be pruned (target components) as an integrated target. 
Therefore, while AAP aims to prune the model to a certain level, the pruning ratio may not be consistent for each target component.

In Table~\ref{tab:sota}, we observe that ICE-Pruning has better accuracy and time consumption in most cases. 
Specifically, at 60\% pruning, ICE-Pruning has up to 7.8\% higher accuracy and is up to $2.96\times$ faster; 
at 70\% pruning, ICE-Pruning has up to 1.35\% higher accuracy and is up to $4.12\times$ faster; 
at 80\% pruning, ICE-Pruning has up to 3.16\% higher accuracy and is up to $9.61\times$ faster. 
Even in cases where the accuracies are similar, such as at 80\% pruning with ResNet-152 and CIFAR-10, ICE-Pruning is still up to $2.67\times$ faster. 
This is because i) the models are so over-parameterized that they are robust to pruning and ii) we skip many fine-tuning operations, which are applied unnecessarily often in other pipelines.

Note that, for ICE-Pruning, the magnitude of time consumption varies significantly between the different models and datasets. For example, for ResNet-152, DenseNet-121 and WRN-101-2 with 60\% pruning, ICE-Pruning takes 1,943, 409, and 6,530 seconds respectively. This is due to the different sizes of models and datasets. Generally, the larger the sizes are, the slower ICE-Pruning will be.

\section{Conclusion}
\label{sec:conclusion}

We have proposed ICE-Pruning, a simple yet effective threshold-guided automatic pruning pipeline based on iterative pruning. ICE-Pruning is based on three key components: a mechanism to determine when fine-tuning is necessary based on an accuracy drop threshold, a novel layer freezing strategy, and a pruning-aware learning rate scheduler. 
This design enables ICE-Pruning to be a plug-and-play solution that supports widely-used pruning criteria. 
Additionally, ICE-Pruning automatically determines the best hyperparameters, which saves the user from the effort of manual tuning. 
In our experiments, we show how different pruning criteria work with ICE-Pruning, and we have explored the impact of different components. 
Finally, we show that ICE-Pruning can maintain accuracy while reducing pruning time significantly, up to $9.61\times$ compared with existing pruning pipelines. 

As future work, we plan to use ICE-Pruning with a broader range of real industrial development environment, models, datasets, and pruning methods to explore its benefits in real-world scenarios. For example, we could use ICE-Pruning to prune large language models (LLMs)~\cite{ye2023comprehensive} for efficient deployment. As another example, we could deploy ICE-Pruning on different types of machine learning servers with diverse hardware configurations to evaluate the impact of hardware platforms on its efficiency.

%Fine-tuning, as a smaller scale of training, can benefit more with larger models, datasets, or engineering scenarios. For a training example, 90-epoch ImageNet-1k training with ResNet-50 on NVIDIA M40 GPU can take 14 days~\cite{2018imagenet}. Using ICE-Axe, this astounding time consumption which may also happen during fine-tuning could be reduced. 
%\pmh{1. no-one is using M40 anymore; 2. why is this relevant -- can your method also accelerate pretraining?}
%\color{green}
%the pertaining uses the same training data in fine-tuning. So a long pre-training could mean a long fine-tuning. I have added some sentences to enhance that.
%\color{blue}
%With ICE-Axe, we can save computing power while keeping the total pruning time low. The potential of applying ICE-Axe in larger-scale industry backgrounds is promising. Fine-tuning, as a smaller scale of training, in pruning DNN models can be very resource-consuming. For example, in the DAWNBench~\cite{coleman2017dawnbench}, the training time scales almost linearly with the inverse number of GPUs. 
%\pmh{again how is this relevant?}
%\color{teal}
%I have swapped the example and the computing power part. The example is just for endorsing computing power saving.
%\color{black}

\balance

\bibliographystyle{IEEEtran}
\bibliography{main}

\end{document}